\documentclass[11pt,a4paper]{article}
\usepackage[hyperref]{acl2021}
\usepackage{times}
\usepackage{latexsym}

\usepackage{microtype}
\usepackage{xcolor}
\usepackage{amsmath}
\usepackage{booktabs}
\usepackage{graphicx}
\graphicspath{ {./figures/} }

\aclfinalcopy 

\title{Lone Pine at SemEval-2021 Task 5: Fine-Grained Detection of Hate Speech Using BERToxic}

\author{Yakoob Khan, Weicheng Ma, Soroush Vosoughi \\
        Department of Computer Science, Dartmouth College \\
        \texttt{\{yakoob.khan.21,weicheng.ma.gr,soroush.vosoughi\}@dartmouth.edu}}

\begin{document}
\maketitle
\begin{abstract}
This paper describes our approach to the \textit{Toxic Spans Detection} problem (SemEval-2021 Task 5). We propose \textbf{BERToxic}, a system that fine-tunes a pre-trained BERT model to locate toxic text spans in a given text and utilizes additional post-processing steps to refine the boundaries. The post-processing steps involve (1) labeling character offsets between consecutive toxic tokens as toxic and (2) assigning a toxic label to words that have at least one token labeled as toxic. Through experiments, we show that these two post-processing steps improve the performance of our model by 4.16\% on the test set. We also studied the effects of data augmentation and ensemble modeling strategies on our system. Our system significantly outperformed the provided baseline and achieved an F1-score of 0.683, placing Lone Pine in the 17\textsuperscript{th} place out of 91 teams in the competition. Our code is made available at \url{https://github.com/Yakoob-Khan/Toxic-Spans-Detection}
\end{abstract}

\section{Introduction}

The promotion of respectful discourse has always been a core tenet of civilized societies. The Cambridge dictionary defines hate speech as ``public speech that expresses hate or encourages violence towards a person or group based on something such as race, religion, sex, or sexual orientation." Online platforms enable malicious actors to hide behind a cloak of anonymity and surreptitiously post toxic comments that are a ``menace to democratic values, social stability and peace" (United Nations). To combat this problem, such platforms often employ human moderators to address offensive content that goes against community standards. However, moderators are unable to manually keep pace with the large volume of user-generated content today. This motivates the development of natural language processing systems to automatically detect hate speech and ensure that online platforms remain healthy and inclusive for all.

There has been extensive research on hate speech detection, with the creation of large datasets ~\citep{Wulczyn:17} and the use of pre-trained text representations ~\citep{Devlin:19} for varied modeling approaches. Competitions on offensive language identification ~\citep{Zampieri:20} have further attracted attention to this topic. Prior work has hitherto focused on classification at the document-level based on various taxonomies, such as whether a given text contains offensive language or if it is targeted towards an individual or group. This line of inquiry does not identify the toxic spans that ascribe a text as hate speech. Doing so will assist human moderators to efficiently locate offensive content in long posts and elucidate further insight into hate speech explainability. 

This motivated the \textit{Toxic Spans Detection} task ~\citep{Pavlopoulos:21} where systems are asked to extract the list of toxic spans that attribute to a text's toxicity. Consider the following example\footnote{We caution readers that the examples included in this work contain explicit language to illustrate the severity and challenges of hate speech detection.}: 

\begin{table}[h]
    \centering
    \begin{tabular}{p{0.1\linewidth}p{0.76\linewidth}}
    \toprule
    \textbf{Text}   & Because he's a {\color{red}\textbf{moron}} and {\color{red}\textbf{bigot}}. It's not any more complicated than that. \\ \midrule
    \textbf{Span}   & [15, 16, 17, 18, 19, 27, 28, 29, 30, 31]  \\ 
    \bottomrule           
    \end{tabular}
    \caption{A sample example from the task dataset.}
\end{table}

As there are two toxic spans in the above text, systems are tasked to extract the character offsets (zero-indexed) corresponding to the sequence of toxic words. This is a challenging task as classification at the word-level is inherently more difficult than at the document-level. The intentional obfuscation of toxic words, use of sarcasm and the subjective nature of hate speech further adds complexity to the problem.

\paragraph{}Our contributions to the task is threefold:
\begin{enumerate}
\item We propose \textbf{BERToxic}, a system that finetunes a pre-trained BERT model with additional post-processing steps to achieve an F1-score of $0.683$, placing Lone Pine in the $17$\textsuperscript{th} place out of 91 teams in the competition.
\item We study the effects of simple data augmentation strategies on our system and find that they yield no improvement in classification performance.
\item We examine late fusion and multi-task learning neural architectures and conclude that they under-perform compared to the standalone BERT model for this task.
\end{enumerate}

\section{Our Approach}

\subsection{Baselines}
To have a better sense of our final system's performance, we initially examined two baseline models. First, we created a trivial model that randomly predicts each character offset of a text as toxic if its $\rho > 0.5$, drawn from a continuous uniform probability distribution. 

To have a stronger baseline model, we fine-tuned the off-the-shelf spaCy NER model provided by the task organizers. This model consists of a multi-hash embedding layer (feed-forward sub-network) that uses sub-word features and an encoding layer consisting of a CNN and a layer-normalized maxout activation function. The model uses a transition-based algorithm that assumes that the ``most decisive information" regarding the entities ``will be close to their initial tokens", with a loss function that optimizes for whole-entity accuracy.  

\subsection{BERToxic}
We framed the toxic spans detection task as a sequence labeling problem and leverage the \textbf{B}idirectional \textbf{E}ncoder \textbf{R}epresentation from \textbf{T}ransformers ~\citep{Devlin:19} model to extract rich feature representations from the input texts. 

\begin{figure}[h]
    \centering
    \includegraphics[width=\columnwidth]{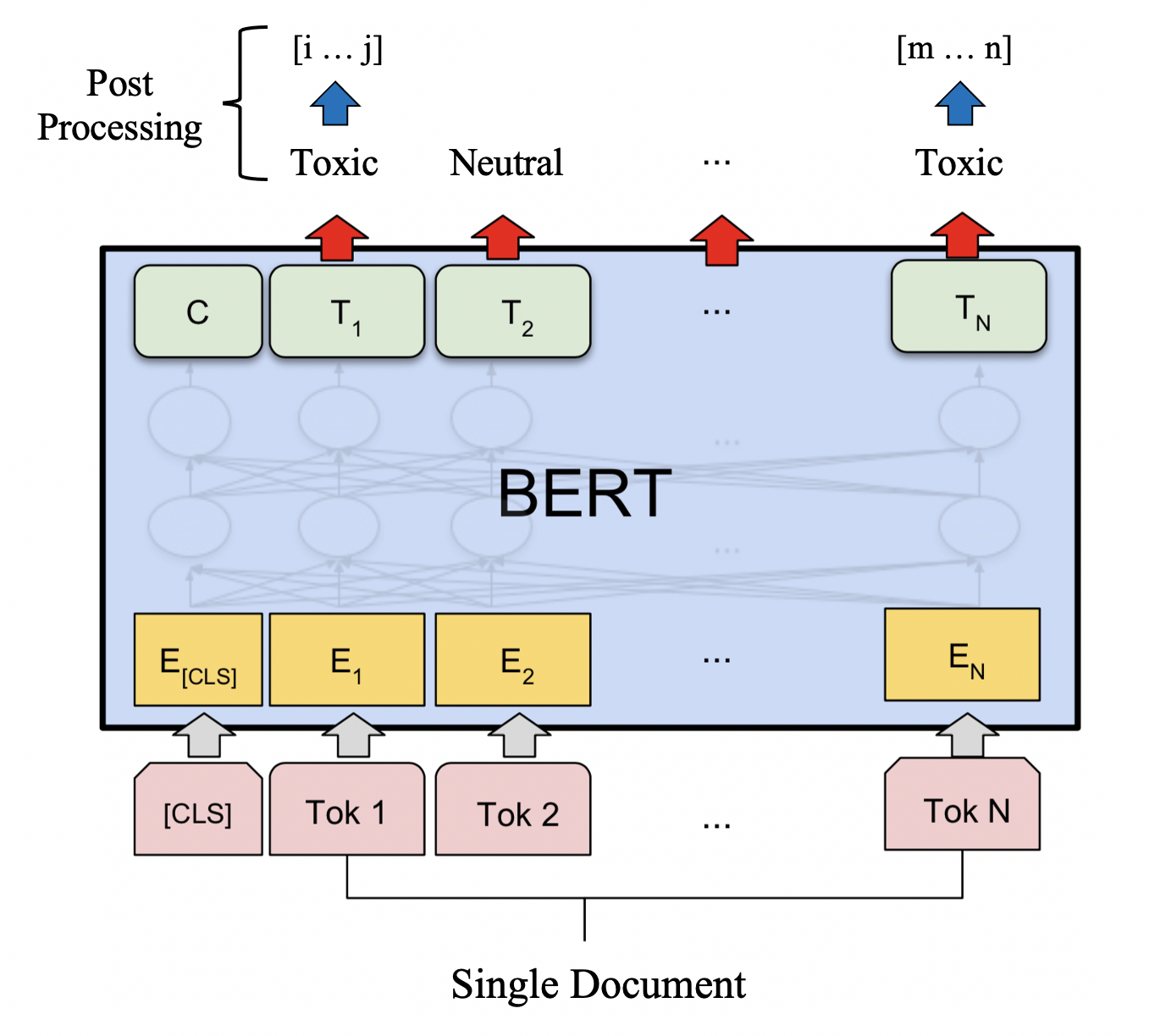}
    \caption{The BERToxic model architecture. Image modified from ~\citep{Devlin:19}.}
    \label{figure:bert}
\end{figure}

The first step in the BERToxic system pipeline (Figure \ref{figure:bert}) was to tokenize the text inputs and generate the word embeddings using BERT's WordPiece tokenizer. This sub-word tokenization algorithm ~\citep{Schuster:12} tokenizes a word like \texttt{"moron"} into \texttt{["mo","\#\#ron"]} and we ensured that the ground truth labels were preserved across all tokens of a word. As BERT uses absolute position embeddings, we padded shorter sequences with \texttt{[PAD]} tokens on the right side such that all tensor inputs are set to equal the maximum sequence length observed for batched parallelized training. Long sequences were truncated to $512$ tokens, the maximum sequence length allowed by BERT. As the data was obtained from online comments that are generally shorter in nature, the truncation procedure was not needed in this task but nevertheless served to handle long sequences if present. We also stored the mapping $$\mathcal{M}: t_{i} \mapsto {(start_{i}, end_{i})}$$ of each token to its relative character offsets in the original string, used for outputting the toxic span predictions at the post-processing stage. 

We performed all of our experiments using the BERT$_\mathrm{BASE}$ model architecture that consisted of $12$ layers, $768$ hidden size, $12$ self-attention heads and $109M$ parameters. The BERT$_\mathrm{LARGE}$ model was not explored in this work due to its compute-intensive nature. Our intuition suggested that letter casing could be helpful for this task as proper nouns (e.g \textit{Muslim}) can be used offensively, so we selected the cased model for our experiments. A token classification head containing a linear layer was applied on top of the final hidden-states output, with a label prediction of $1$ denoting a toxic token, $0$ otherwise. For each token $t_{i}$ labeled as toxic, we utilized $\mathcal{M}$ to output all character indices in the range of ${(start_i, end_i)}$ inclusive as the toxic span of this token.

Additionally, our system performed two post-processing steps to refine the boundary predictions. Consider the following tokenized sequence: $$t_1, \cdots, t_i, t_{i+1}, t_{i+2}, \cdots, t_n$$ First, for any two consecutive tokens $t_i$ and $t_{i+1}$ whose prediction labels are toxic, we output the character indices in the range of $(end_{i} + 1, start_{i+1} - 1)$ inclusive as toxic as well. This had the effect of including the delimiter characters between consecutive toxic words, thereby detecting toxic phrases. Second, recall that BERT's WordPiece tokenizer could split a word into multiple tokens, say $t_i, t_{i+1}$ and $t_{i+2}$. If at least one token was predicted toxic by the model, our system assigned a toxic label to all constituent tokens of this word. This achieved coherence in the prediction of toxic words and phrases, thus avoiding incomplete word piece issues. 

We also attempted to vary the thresholds of the confidence scores before SoftMax for toxic token predictions but observed no improvement in performance.

\subsection{Data Augmentation}
Data augmentation is widely used to improve the generalization of models by acting as a regularizer to reduce overfitting. While various sophisticated techniques exist to artificially enhance the size and quality of the training set without collecting additional manually labeled examples, we chose to apply the set of \textbf{E}asy \textbf{D}ata \textbf{A}ugmentation (EDA) techniques ~\citep{Jason:19} to generate synthetic training data for this task.

The four operations in EDA are Synonym Replacement (SR) using WordNet \citep{Miller:95}, Random Insertion (RI), Random Swap (RS) and Random Deletion (RD) of words in a document. Shorter documents are disproportionately more affected by these operations if a fixed number of words are modified per document. To ensure that all documents experienced the augmentation strength proportionately, the number of words $n$ modified was varied based on the document length $l$ using the formula $ n = \alpha \cdot l $, where $\alpha$ is a hyper-parameter that indicates the percentage of words changed per document. Each operation was applied once per document and care was taken to ensure that the ground truth labels were preserved.

Our experiments revealed that the recommended value of $\alpha = 0.1$ was too low for this task and we observed small but consistent improvements as $\alpha$ increases. Furthermore, we noted that the SR technique alone leads to better performance than using all four operations to create the augmented training set for this task.

We also attempted data augmentation using an external dataset, HateXplain ~\citep{Mathew:20}, that contains $20,148$ documents with word-level annotations that we processed to conform to this task's data format. Each document consisted of $2$ - $3$ annotations and we used their intersection to maximize the inter-annotator agreement in constructing the ground truth labels. HateXplain's annotation strategy appeared to be different and included labeling pronouns, conjunctions and stop words as toxic when located between offensive words. We removed such toxic labels so that the external dataset annotation was more similar to this task. When our task dataset was augmented with the full external dataset, the model experienced underfitting, while removing all the non-toxic labeled documents from the external dataset alleviated the issue to some extent. 

\subsection{Ensemble Modeling}
Ensemble modeling is an approach where multiple different models are trained and their predictions are aggregated. By adding bias to counter the variance of a single model, this line of work has been shown to improve the predictive performance of a system \citep{Liu:19}. While numerous ensemble modeling techniques like boosting, bagging, etc. exist, we investigated two techniques of interest: late fusion and multi-task learning.

We reframed the problem as a binary classification task and trained a sequence classifier to predict whether a given sentence is toxic. In the late fusion approach, we utilized NLTK's tokenizer to split each document into sentences. If a sentence contained a ground truth toxic span, we assigned the toxic class label $1$, $0$ otherwise. In this way, a binary classification dataset was created to separately fine-tune a pre-trained BERT sequence classifier. We hypothesized that token labels should be predicted toxic only if the corresponding sentence was classified as toxic as well. Late fusion was performed at the prediction phase, where both the sequence and token classifiers voted in the predictions by having the former model filter toxic sentences on which the latter model made final toxic span predictions. 

Rather than fine-tuning the two models separately, we also investigated if multi-task learning (MTL) improved the predictive performance of the ensemble model. We hypothesized that a training regime where the two classifiers were learned jointly could be useful as the knowledge gained in learning one task could benefit the other. To perform MTL, we fine-tuned an MT-DNN ~\citep{Liu:19} model where the text encoding lower BERT layers are shared across the two tasks while the top layers are task-specific. 

\section{Experiments}
The following sections describe the experimental set-up of our work.

\subsection{Dataset}
The task data was sourced from the Civil Comments dataset ~\citep{Borkan:19}, which contains public comments made between $2015$ - $2017$ that appeared on approximately $50$ English-language news sites across the world. As the original dataset contained only document-level class labels, the task organizers selected a subset of the data for crowd-sourced toxic spans annotation. For the data split, we chose to fine-tune our models using the entire provided training dataset $(N = 7939)$ to maximize performance, validate using the trial dataset $(N = 690)$, and submit our predictions using the test data $(N = 2000)$. The test labels were withheld during the evaluation phase of the competition and were only released afterward.

\subsection{Evaluation Metric}
To evaluate the performance of the models, the task organizers employed a variant of the F1-score \citep{Martino:19}. For a document $d$, define $S_d$ as the set of toxic character offsets predicted by a system and $G_d$ as the set of ground truth annotations. Then the F1-score of the system with respect to ground truth $G$ for $d$ is defined as 
$${F_1}^d(G) = \frac{2 \cdot P^d(G) \cdot R^d(G)}{P^d(G) + R^d(G)}$$ 
where 
$$P^d(G) = \frac{|S_d \cap G_d|}{|S_d|}$$ 
$$R^d(G) = \frac{|S_d \cap G_d|}{|G_d|}$$ 
If a document has no ground truth annotation $(G_d = \emptyset)$, or the system outputs no character offset prediction $(S_d = \emptyset)$, we set  
\[ {F_1}^d(G) = \begin{cases} 
              1 & G_d = S_d = \emptyset \\
              0 & otherwise
           \end{cases}
\]
We finally take the arithmetic mean of ${F_1}^d(G)$ over all the documents of an evaluation dataset to obtain a single F1-score for the system.

\subsection{Implementation Details}
We utilized the PyTorch framework for the development of our system, HuggingFace's transformers library \cite{HuggingFace} for the BERT-based models and Microsoft's implementation of the MT-DNN model. All models were trained on Google Colab Pro's High-RAM environment using a single NVIDIA P100 GPU. The training policy used the following hyper-parameters: batch size of $16$, sequence length of $512$, weight decay of $0.01$. For optimization, we used Adam with a learning rate of 5e-5 and a linear warm-up schedule over $500$ steps. Except for MT-DNN\footnote{MT-DNN was fined-tuned for $3$ epochs with a batch size of $8$.}, all our models were fine-tuned for approximately $2$ epochs and we practiced early stopping by monitoring the dev F1-score to reduce overfitting. The EDA experiment was performed with $\alpha = 0.8$ using only the SR technique. All other hyper-parameters were set to their default values according to HuggingFace's implementation. We set a random seed for all our experiments and open-sourced the code for reproducibility. 

\section{Results}

\begin{table*}[]
\centering
\begin{tabular}{lllllll}
\toprule
\textbf{Model}              & \multicolumn{3}{c}{\textbf{Dev}} & \multicolumn{3}{c}{\textbf{Test}}   \\ \hline
                            & Precision & Recall &    F1    & Precision  & Recall  &        F1       \\ \hline
        Random              &   0.143   &  0.463 &  0.175   &    0.089   &  0.413  &      0.122      \\ \hline
        SpaCy               &   0.692   &  0.588 &  0.595   &    0.664   &  0.686  &      0.656      \\ \hline
        BERToxic            &   0.781   &  0.678 &  0.681   &    0.683   &  0.732  &  \textbf{0.683} \\ 
        + EDA               &   0.787   &  0.683 &  0.684   &    0.681   &  0.725  &      0.678      \\
        + HateXplain        &   0.792   &  0.674 &  0.681   &    0.683   &  0.721  &      0.678      \\ \hline
        BERT late fusion    &   0.733   &  0.636 &  0.639   &    0.675   &  0.709  &      0.669      \\ \hline
        BERT multi-task     &   0.744   &  0.629 &  0.634   &    0.665   &  0.694  &      0.656      \\ \bottomrule
\end{tabular}
\caption{A summary of the performance of all our models, reporting the precision and recall scores along with the F1 evaluation metric used for the competition. The BERToxic model outperformed the strong spaCy baseline by $4.16$\% on the test set, placing Lone Pine in the $17$\textsuperscript{th} place out of 91 teams. In comparison, the top-ranked submission achieved an F1-score of 0.708. The experiments revealed that our data augmentation and ensemble modeling strategies did not outperform the standalone BERT model.}
\label{table:summary}
\end{table*}

\begin{figure}[h]
    \centering
    \includegraphics[width=\columnwidth]{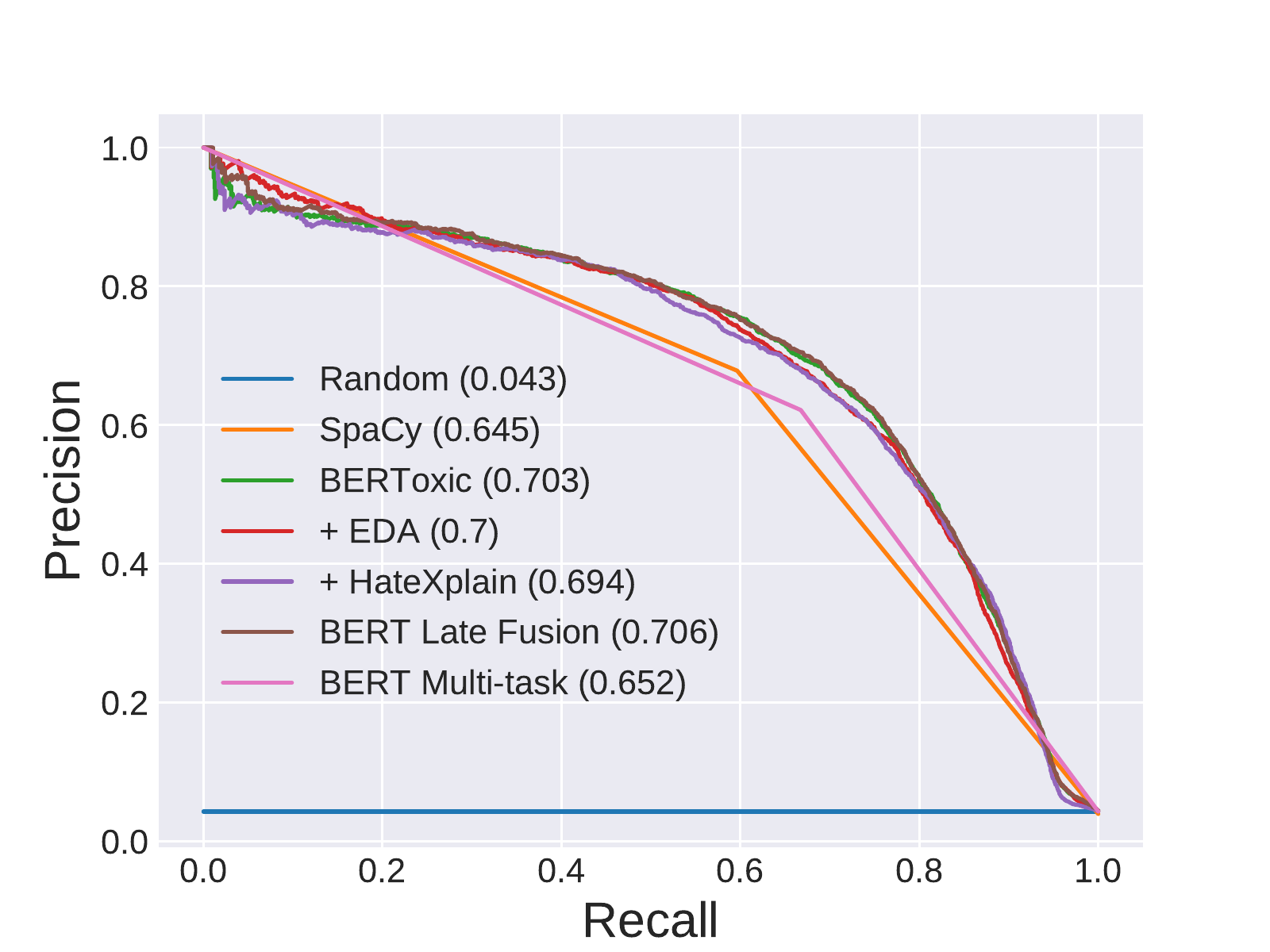}
    \caption{Comparison of the precision-recall curves of all the models at the token level on the test set. The area under the curve is enclosed within parentheses.}
    \label{figure:precision-recall}
\end{figure}

\begin{figure}[h!]
    \centering
    \includegraphics[width=\columnwidth]{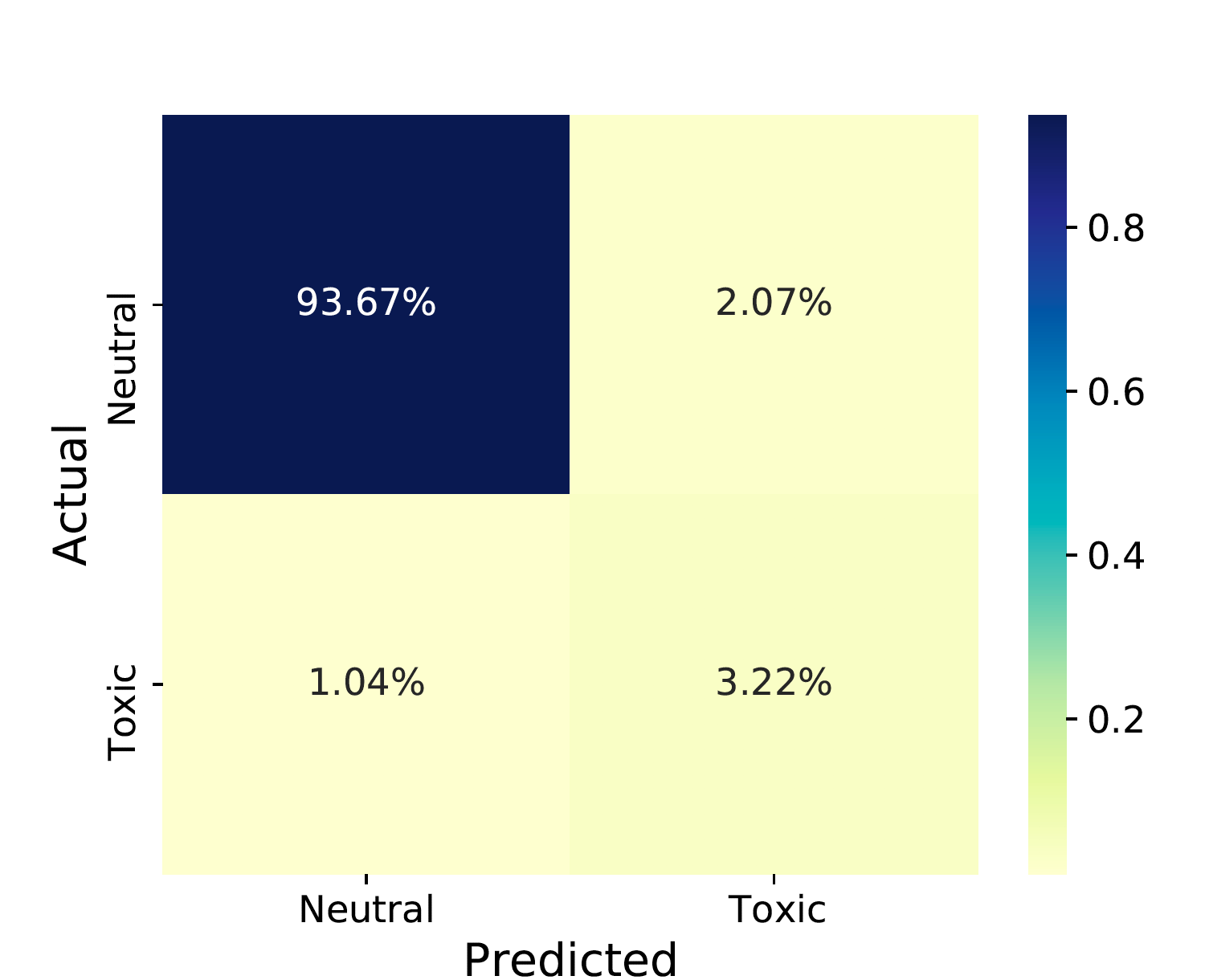}
    \caption{A confusion matrix of the BERToxic system at the token level, revealing insights about the classification performance in each category and highlighting the imbalance of the class labels in the test set.}
     \label{figure:confusion-matrix}
\end{figure}

On the following page, Figure \ref{figure:precision-recall} visualizes the precision-recall curves\footnote{The curves for the spaCy and BERT multi-task model are less detailed due to the ambiguity in obtaining the probability scores from their respective implementations, necessitating the use of their predicted labels instead.} of all the models and Table \ref{table:summary} summarizes their performance metrics. Figure \ref{figure:confusion-matrix} shows the confusion matrix of our best performing BERToxic model and Table \ref{table:predictions} highlights selected predictions that it made.

An interesting observation we noted from Table \ref{table:summary} was that the F1 scores for the test set were higher than the dev set for many of the models. We hypothesize that this is because the models have an inductive bias to predict shorter toxic spans, evidenced by the average ground truth span length of $7.2$ in the test set and $14.7$ in the dev set. 

Our proposed system performed well at the toxic spans detection task, showing strength in identifying profanity and common toxic words like ``idiot" and ``stupid". The model identified the obfuscation of offensive words and successfully detected hate speech from such adversarial cases (Example 1).

\begin{table}[h]
    \centering
    \begin{tabular}{@{}ll@{}}
    \toprule
    
        1.  & {\color{red}Kill} this \textit{{{\color{red}F'n W*ore}}} on site.  \\ \midrule
        2.  & .. how I am an {\color{red}ignorant fool} ..         \\ \midrule
        3.  & {\color{red}Nazi boneheads} deserve being {\color{red}punched}.   \\ \midrule
        4.  & @ remoore {\color{red}Shut up}, \textit{{\color{red}racist}}. \\ \midrule
        5.  & \textit{Cruz is a piece of {\color{red}garbage} a globalist fraud} \\ 
    \bottomrule           
    \end{tabular}
    \caption{Selected examples obtained from the test set. BERToxic's predictions are shown in {\color{red}red} while ground truth annotations are \textit{italicized}.}
    \label{table:predictions}
\end{table}

The error analysis revealed that the system lacked nuance as it would sometimes classify toxic words used in neutral contexts (Example 2). It is also worth mentioning that there was considerable noise in the ground truth annotations. Our manual inspections concurred with the model's predictions that some words and phrases were used in offensive contexts but the annotators thought they were neutral (Example 3 and 4). Furthermore, we observed some inconsistencies in the labeling scheme as some annotations spanned entire sentences (Example 5) while others only highlighted a few words in the sentence. These issues point to the subjective nature of hate speech and the challenges involved in its fine-grained classification.

We found through our ablation studies of data augmentation that generating synthetic data using the EDA techniques did not improve the performance of the system. This suggested that the dataset size does not appear to be the limiting factor affecting the performance of BERT in this task. Using HateXplain's external dataset, we learned that different data sources and annotation guidelines can introduce noise that hurts the performance of models.

Finally, the ensemble modeling strategies we explored did not outperform the standalone BERT model. The late fusion technique performed slightly better than the spaCy baseline, but it seemed that the sequence classifier made errors on similar parts of the input space as the token classifier. The multi-task learning approach under-performed compared to late fusion, suggesting that the sequence labeling and classification tasks are not closely related enough to benefit their joint training.

\section{Conclusion and Future Work}
In this work, we have proposed BERToxic, an empirically powerful system that performed fine-grained detection of hate speech. We found that our exploration of data augmentation and ensemble modeling strategies did not outperform the standalone model. The error analysis revealed that BERT lacked nuance in understanding the use of offensive words in neutral contexts and encountered boundary detection issues when faced with noisy ground truth annotations. 

Future avenues of work could address these limitations and explore other transformer-based models to develop more robust hate speech detectors. We hope that our findings inspire more creative approaches towards fine-grained detection of hate speech so that online discourse can remain healthy and inclusive for all.

\section*{Acknowledgments}
Yakoob Khan is thankful to be supported by the Stamps Scholarship funded by Dartmouth College and the Strive Foundation.  

\bibliographystyle{acl_natbib}
\bibliography{anthology,acl2021}


\end{document}